\documentclass[10pt,twocolumn,letterpaper]{article}
\usepackage{booktabs}
\usepackage{makecell}
\usepackage{wacv}              % To produce the CAMERA-READY version
\definecolor{wacvblue}{rgb}{0.21,0.49,0.74}
\usepackage[pagebackref,breaklinks,colorlinks,allcolors=wacvblue]{hyperref}
\usepackage{booktabs}

\def\wacvPaperID{3041} % *** Enter the WACV Paper ID here
\def\confName{WACV}
\def\confYear{2026}

\usepackage[table]{xcolor}
\usepackage{adjustbox}

\usepackage{algorithm}
\usepackage{algorithmic}

\title{
Fast 2DGS: Efficient Image Representation with Deep Gaussian Prior
}

\author{Hao Wang$^1$\thanks{Corresponding author}, Ashish Bastola$^1$, Chaoyi Zhou$^1$, Wenhui Zhu$^2$, Xiwen Chen$^1$, Xuanzhao Dong$^2$, \\Siyu Huang$^1$, and Abolfazl Razi$^1$\\
$^1$Clemson University\\
$^2$Arizona State University\\
{\tt\small \{hao9, abastol, chaoyiz, xiwenc, siyuh, arazi\}@clemson.edu, \{wzhu59, xdong64\}@asu.edu}\\
}
\begin{document}
\maketitle

\begin{abstract}
% Image representation is a fundamental cornerstone of computer vision, serving as the bedrock for compression, editing, and transmission tasks. While Implicit Neural Representations (INRs) have demonstrated impressive compactness and resolution independence, they fundamentally suffer from slow per-image optimization (often taking minutes) and a "black-box" nature that hinders interpretability. 

As generative models become increasingly capable of producing high-fidelity visual content, the demand for efficient, interpretable, and editable image representations has grown substantially. 
Recent advances in 2D Gaussian Splatting (2DGS) have emerged as a promising solution, offering explicit control, high interpretability, and real-time rendering capabilities ($>1000$ FPS). 
However, high-quality 2DGS typically requires post-optimization. Existing methods adopt random or heuristics (e.g., gradient maps), which are often insensitive to image complexity and lead to slow convergence ($>10s$). More recent approaches introduce learnable networks to predict initial Gaussian configurations, but at the cost of increased computational and architectural complexity.
To bridge this gap, we present \textbf{Fast-2DGS}, a lightweight framework for efficient Gaussian image representation. Specifically, we introduce Deep Gaussian Prior, implemented as a conditional network to capture the spatial distribution of Gaussian primitives under different complexities. In addition, we propose an attribute regression network to predict dense Gaussian properties.
Experiments demonstrate that this disentangled architecture achieves high-quality reconstruction in a single forward pass, followed by minimal fine-tuning. More importantly, our approach significantly reduces computational cost without compromising visual quality, bringing 2DGS closer to industry-ready deployment.
Our source code and sample datasets are available at: 
\url{https://github.com/Aztech-Lab/Fast-2DGS}

% state-of-the-art trade-offs between quality and speed, matching the fidelity of GaussianImage while reducing total optimization time by over $10\times$ and model size by $40\times$ compared to concurrent network-based approaches.
\vspace{-0.5cm}
\end{abstract}

%-------------------------------------------------------------------------
\section{Introduction}
\label{sec:intro}

The exponential growth of visual data in the era of Artificial Intelligence Generated Content (AIGC) has reignited the demand for efficient, scalable, and manipulable image representations \cite{MiraGe}. 
In domains such as digital entertainment\cite{MEMCNet}, content creation\cite{Freestyle,SODA,zhao2024uni}, and gaming\cite{VLMGaming}, visual data plays a critical role in spatial perception and reasoning\cite{GaussianToken}. 
% As generative models increasingly integrate visual modalities, the ability to represent and manipulate such information becomes a core requirement for advancing AIGC technologies since images encode rich spatial structures and semantic cues that are essential for both human understanding and machine intelligence. 
Traditional pixel-based formats (JPEG, PNG)\cite{RGBNoMore}, while ubiquitous, suffer from spatial redundancy and lack semantic structure \cite{ImageGS}. 
In contrast, neural representations aim to compress visual signals into deep network parameters or compact primitives \cite{INRCompression}.
% Implicit Neural Representations (INRs) \cite{sitzmann2020implicit, chen2021learning} have pioneered this shift by modeling images as continuous functions parameterized by Multi-Layer Perceptrons (MLPs). 
While effective for compression, Implicit Neural Representations (INRs) face significant latency issues \cite{InstantNGP}.
% : fitting a high-resolution image requires solving a non-convex optimization problem from scratch, often taking minutes per image. 
% Grid-based accelerations like I-NGP \cite{mueller2022instant} mitigate this but introduce prohibitive memory footprints (hundreds of MBs), rendering them unsuitable for edge devices.

\begin{figure}
    \centering
    \includegraphics[width=1\linewidth]{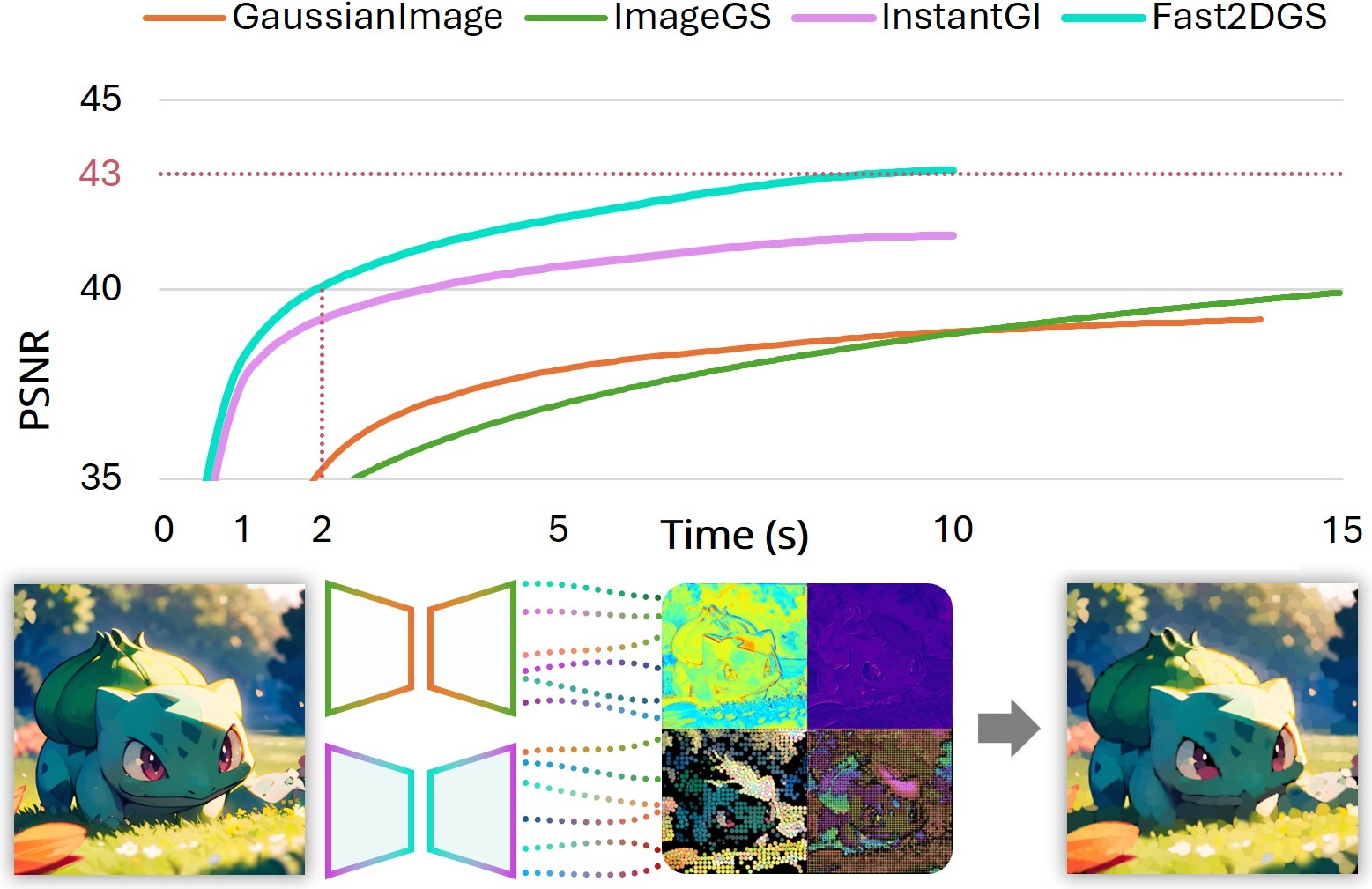}
    \caption{Concept of the proposed Fast 2DGS. We adopt two separate lightweight networks to perform prediction of Gaussian primitives. Our method achieves 40 PSNR in 2 seconds of fine-tuning, and reaches 43 within 10 seconds.}
    \vspace{-0.5cm}
    \label{fig:cover}
\end{figure}

Recently, 2D Gaussian Splatting (2DGS) \cite{GaussianImage} has been proposed as an explicit alternative. By representing an image as a collection of 2D anisotropic Gaussians explicitly, 2DGS offers fast decoding ($>1000$ FPS) using efficient rasterization, and intuitive editability \cite{GaussianImage,Neural2DGSVideo,Large2DGS}. 
However, the quality of 2DGS is highly dependent on initialization. For instance, random initialization places primitives uniformly. This forces the optimizer to slowly migrate points to high-frequency regions, requiring $\sim$10k iterations to converge. 
In contrast, heuristic-based initialization utilizes gradient magnitude or saliency maps \cite{ImageGS}, while offering improvements over random initialization, these static heuristics are limited in dynamic scenarios\cite{GaussianImage,ImageGS}. 
% For instance, they fail to differentiate between simple and complex scenes, or between low- and high-resolution images, resulting in suboptimal prioritization of image regions during representation learning \cite{GaussianImage,ImageGS}.

% while better, these static maps ignore the capacity of the representation. These methods produce static importance maps that do not adapt to the varying complexity or resolution of images. For instance, a low-capacity representation (small $K$) would benefit from focusing on global structures, whereas a high-capacity model (large $K$) can afford to capture fine-grained texture details. However, static heuristics fail to account for such variations, leading to suboptimal allocation of representational resources across different image types and resolutions.
% For example, a low-capacity model (small $K$) should focus on global structure, while a high-capacity model (large $K$) should focus on texture details. Static heuristics fail to capture this distribution shift.

Instant GaussianImage \cite{InstantGI} introduced a network-based initialization strategy customized for the 2D Gaussian Splatting task. 
Instead of relying on iterative optimization from scratch, it employs a neural network as an encoder to predict Gaussian parameters, including spatial locations, covariance structures, and color attributes from the input image. 
The predicted probability maps (PPM) are converted into discrete Gaussian centers using a dithering-based sampling algorithm\cite{FloydDithering}, enabling adaptive and structure-aware initialization. 
This hybrid strategy significantly accelerates convergence, reducing encoding time while maintaining high reconstruction quality from the outset. 
Nevertheless, the design of Instant GaussianImage introduces several practical concerns. Its reliance on feature engineering complicates the Gaussian encoding process, and the heavy network architecture introduces computational overhead. 
% More critically, its discretization step breaks gradient flow and fails to fully exploit the benefits of discrete optimization. 
% Subsequent enhancements also lean heavily on empirical engineering choices rather than learned priors. 
While it performs well in experiments, the increasing complexity and computational cost pose limitations for scalable deployment in industrial applications.

To address these challenges, we present \textbf{Fast2DGS}, a framework that transforms Gaussian initialization from a heuristic guess into a learning-based task. 
Specifically, we introduce Deep Gaussian Prior, where the spatial distribution of Gaussians is learned through an iterative bootstrapping loop.
% iterative sampling-optimization loop. 
% Furthermore, we employ a classical lightweight UNet architecture for Gaussian feature prediction to reduce the rely on manual designed feature engineering.
Furthermore, we employ a lightweight UNet for Gaussian feature prediction. By stripping away the feature engineering pipelines, we demonstrate that effective Gaussian attribute prediction does not necessitate elaborate architectural designs, as show in Figure \ref{fig:cover}. 
% This choice validates that our proposed iterative prior learning mechanism is robust enough to drive convergence end-to-end using a standard backbone.
In summary, our contribution is three-fold:
\begin{itemize}
\item We propose Deep Gaussian Prior, an initialization strategy learned through an iterative optimization-sampling loop. By simulating the optimization trajectory, our method captures a content-aware distribution that breaks the uniform bias of random initialization, significantly accelerating convergence.
\item We present a streamlined framework that anchors Gaussian cardinality ($K$) to compression rates. By employing a lightweight backbone without complex feature engineering, we transform the ill-posed Gaussian allocation into a tractable, batch-parallelizable learning task.
\item We demonstrate that our framework achieves superior trade-offs between reconstruction quality, inference latency, and cross-dataset generalizability compared to existing Gaussian image approaches.
\end{itemize}

\section{Related Work}
\label{sec:related}

\subsection{Image Representation}
Image representation serves as the foundation across various computer vision and signal processing tasks, including image compression\cite{INRCompression}, super-resolution\cite{GaussianSR,DRLN}, and generative modeling\cite{ColdDiffusion,Freestyle,SODA}. 
Traditionally, images have been represented via discrete grid-based formats (e.g., pixels) or transform-domain coefficients (e.g., DCT\cite{DCT,DCTSR}, Wavelets\cite{RobustFormer}).
In recent years, Implicit Neural Representations (INRs) have emerged as dominant the deep learning domain by utilizing multilayer perceptrons (MLPs) to parameterize images as continuous functions mapping spatial coordinates to color values \cite{INRCompression}. 
INR models such as SIREN\cite{SIREN}, WIRE\cite{WIRE}, and COIN++\cite{COIN_++}  have demonstrated remarkable capabilities in capturing high-frequency details and achieving resolution independence.
However, the implicit nature of these methods imposes significant computational bottlenecks \cite{InstantNGP}. 
For instance, decoding a single pixel requires a full forward pass through the neural network, resulting in slow inference speeds and substantial GPU memory consumption. 
% Although feature grid-based methods (e.g., Instant-NGP) have accelerated this process, they often rely on fixed data structures that lack content adaptivity and still demand heavy compute for coordinate querying. 
Furthermore, implicit models typically function as black boxes and lack efficient support for hardware-friendly random access or local editing.

% Driven by the need for computational efficiency, interpretability, and generative controllability, explicit representations achieve faster decoding and better interpretability, facilitating robust feature engineering through geometric priors
% More importantly, the support of local editability is essential for interactive graphics, which aligns with modern generative AI pipelines, offering a scalable, token-based format that bridges the gap between efficient storage and controllable high-fidelity content generation.

\subsection{2D Gaussian Splatting for Images}
Gaussian Splatting (GS) has recently emerged as a powerful tool for 3D scene reconstruction and real-time rendering\cite{SceneSplat,Latent3DGS}. Compared to implicit neural representations that rely on MLPs, 3D GS employs explicit 3D Gaussians as representation units, combined with a differentiable tile-based rasterization pipeline\cite{3DGSEnhancer,WildGS}. 
% In standard 3D workflows, Gaussians are projected onto a 2D plane, sorted based on their depth information, and then rendered using $\alpha$-blending to synthesize novel views\cite{}. 
This approach has demonstrated exceptional performance in tasks ranging from SLAM to dynamic scene modeling \cite{DA3,MotorFocus}.

Directly applying the 3D GS framework to 2D images
% is theoretically possible by fixing the camera view, it 
introduces significant redundancy and computational inefficiencies as
3D Gaussians typically carry up to 59 parameters (e.g., spherical harmonics for view-dependent color) to ensure multi-view consistency \cite{CAT3DGS}, which creates massive storage overhead when representing a single static image \cite{CAT3DGS}. 
% Furthermore, the core mechanism of 3D GS: depth-based sorting, becomes ill-posed in the 2D domain where explicit depth cues are absent[cite: 77]. 
Furthermore, without accurate depth, sorting becomes arbitrary, and the standard $\alpha$-blending may prematurely terminate ray marching, leading to the underutilization of Gaussian primitives \cite{GaussianImage}. 
% Consequently, the transition from 3D to 2D requires shifting the focus from geometric consistency to compact representation and high-fidelity reconstruction with minimal parameters[cite: 80, 81].

% Drawing inspiration from 3D Gaussian Splatting \cite{kerbl3dgaussian}, recent works have adapted the formalism to 2D. 
% While Gaussian Splatting (GS) has achieved transformative success in 3D scene reconstruction and real-time rendering[cite: 35, 109], its direct application to single 2D image representation presents unique challenges. 

To address these challenges, GaussianImage\cite{GaussianImage} introduced a hardware-level optimization for 2D image representation
% Specifically, it proposed replacing the sequential, depth-reliant $\alpha$-blending with an accumulated summation mechanism. 
by removing the need for sorting, allowing for highly parallelized rasterization that achieves decoding speeds of approximately 2000 FPS \cite{GaussianImage}. 
% Furthermore, it integrated vector quantization to construct a functional image codec[cite: 12]. 
However, despite these advancements, significant challenges remain.
Specifically, while the inference speed is ultra-fast, the encoding process (fitting) is still orders of magnitude slower than VAE-based approaches \cite{InstantGI}. 
% Second, although it surpasses implicit methods like COIN, there remains a notable rate-distortion gap compared to state-of-the-art VAE compression methods. Additionally, the standard decomposition of covariance matrices (e.g., rotation-scaling) has been shown to be sensitive to quantization noise, suggesting that specialized compression algorithms for Gaussian attributes are still an open research area[cite: 754, 758].
% To address the training inefficiency of GaussianImage, which relies on a slow two-stage pipeline of overfitting and vector-quantization fine-tuning [cite: 944][cite_start], 
Image-GS \cite{ImageGS} then introduced a content-adaptive method to address this inefficiency. 
Instead of random initialization, it employs gradient-guided sampling to allocate primitives to high-frequency regions and progressively spawns Gaussians in areas with persistent reconstruction errors. While it accelerates the fine-tuning speed, it struggles to model high-frequency pixel-level sensor noise from natural photorealistic images.
% Furthermore, it replaces the global accumulated summation with a tile-based top-K normalization rendering scheme, which not only enables hardware-friendly random access but also accelerates decoding speed by $4\times$[cite: 854, 1036]. [cite_start]
% However, limitations persist in representing natural photorealistic images; the explicit Gaussian primitives struggle to model high-frequency pixel-level sensor noise efficiently, often prioritizing prominent structural features while losing fine-grained details under constrained memory budgets[cite: 1173, 1258].
% To further mitigate the training latency and sampling heuristics of previous methods, 
Instant GaussianImage \cite{InstantGI} proposes a generalizable network framework for rapid Gaussian initialization by introducing a learnable Position Probability Map (PPM) supervised by optimal Gaussian density.
% Additionall, Coupled with Floyd-Steinberg dithering for adaptive spatial distribution[cite: 1572, 1619]. [cite_start]By leveraging Delaunay Triangulation to extract geometric features and predicting opacity rather than direct color to stabilize convergence, 
It achieves high-fidelity reconstruction in a single feedforward pass followed by minimal fine-tuning. However, this sophisticated initialization pipeline requires an intensive triangulation process and additional memory footprint for the feature extraction network.

\begin{figure*}[htbp]
    \centering
    \includegraphics[width=1\linewidth]{Fig/frame.png}
    \caption{Overview of Fast-2DGS. Taking batch images as input, (i) the Position U-Net first predicts a Heatmap representing the Gaussian Prior, from which initial positions are sampled. (ii) These positions are then combined with predicted features (offsets, scales, rotations, colors) to form full Gaussians for image reconstruction and loss computation.}
    \label{fig:frame}
        % \vspace{-0.5cm}
\end{figure*}

\section{Method}
\label{sec:method}

\subsection{Gaussian Image Rendering}
\label{sec:problem}

\noindent\textbf{Gaussian Function.} We represent an image $I \in \mathbb{R}^{H \times W \times 3}$ using a set of $K$ 2D Gaussian primitives $\Theta = \{ \mathbf{g}_k \}_{k=1}^K$. Unlike 3DGS approaches, we employ an optimized 2DGS rendering engine from ImageGS \cite{ImageGS}, where the opacity attribute is omitted as occlusion handling is unnecessary for 2D representation. Thus, each Gaussian primitive is defined by: \textbf{Center position} $\boldsymbol{\mu}_k \in \mathbb{R}^2$ 
% (normalized to $[-1, 1]$)
, \textbf{Covariance matrix} $\boldsymbol{\Sigma}_k \in \mathbb{R}^{2 \times 2}$, and \textbf{Color vector} $\mathbf{c}_k \in \mathbb{R}^3$ (RGB).
The spatial influence of the $k$-th Gaussian at pixel $\mathbf{x}$ is then given by:
\begin{equation}
    G_k(\mathbf{x}) = \exp\left(-\frac{1}{2}(\mathbf{x} - \boldsymbol{\mu}_k)^T \boldsymbol{\Sigma}_k^{-1} (\mathbf{x} - \boldsymbol{\mu}_k)\right).
\end{equation}

\noindent\textbf{Covariance Decomposition.} To maintain the positive semi-definite constraint of $\boldsymbol{\Sigma}_k$, we employ the Rotation-Scaling (RS) decomposition from ImageGS \cite{ImageGS}, as it decouples orientation from extent, which is easier for neural networks to learn:
\begin{equation}
    \boldsymbol{\Sigma}_k = \mathbf{R}(\theta_k) \mathbf{S}(\mathbf{s}_k) \mathbf{S}(\mathbf{s}_k)^T \mathbf{R}(\theta_k)^T,
\end{equation}
where $\mathbf{R}(\theta_k) = \begin{bmatrix} \cos\theta_k & -\sin\theta_k \\ \sin\theta_k & \cos\theta_k \end{bmatrix}$ is the rotation matrix and $\mathbf{S}(\mathbf{s}_k) = \text{diag}(s_{x,k}, s_{y,k})$ is the scaling matrix.

\noindent\textbf{Rendering Model.} We adopt the normalized tile-based rendering model from ImageGS to ensure high-fidelity reconstruction \cite{ImageGS}. Since 2D representation does not require depth sorting, the alpha-blending is replaced with a normalized weighted summation, which effectively regularizes the pixel color reconstruction:
\begin{equation}
    \hat{I}(\mathbf{x}) = \frac{\sum_{k \in \mathcal{N}(\mathbf{x})} G_k(\mathbf{x}) \cdot \mathbf{c}_k}{\sum_{k \in \mathcal{N}(\mathbf{x})} G_k(\mathbf{x}) + \epsilon},
    \label{eq:render}
\end{equation}
where $\mathcal{N}(\mathbf{x})$ denotes the set of relevant Gaussians covering pixel $\mathbf{x}$ (e.g., top-K filtering), and $\epsilon$ ensures numerical stability. 
For convenience, we use 
\begin{equation}
   \hat{I}(\mathbf{x}) = \sum_{k=1}^K \mathcal{R}(\mathbf{g}_k) =  \mathcal{R} (\Theta) 
\end{equation}
to represent the rasterization of $K$ Gaussians.
This formulation remains fully differentiable w.r.t. all parameters $\Theta$.

%-------------------------------------------------------------------------

\subsection{Regularize ill-Posedness via Sampling}
% \section{Method}

Directly predicting the set $\Theta$ from an image $I$ using a neural network is challenging, where a network outputting a fixed-size tensor faces a permutation ambiguity ($K!$ possible orderings) as $\Theta$ is unordered \cite{SetTransformer}.
Furthermore, the inverse problem $\mathcal{R}^{-1}(I)$ is typically ill-posed, as multiple Gaussian configurations can yield identical images (e.g., one large Gaussian vs. many small overlapping ones) \cite{DH1971}.

To regularize the search space of this multi-solution problem, we anchor the decomposition to a hard $K$ (num. of Gaussians) primitives, following the InstantGI\cite{InstantGI}. 
This constrains the search space from a high-dimensional continuous manifold to a structured subspace \cite{NPhard1995,LowRankManifold}.
Rather than deterministic regression, we formulate position initialization as a sampling process from a learned probability distribution, where deep neural networks can handle naturally \cite{Uncertainty,dong2025cunsb}.

However, bridging the continuous spatial distribution (Heatmap $\mathcal{H}$) and the discrete set of Gaussian positions ($\mathbf{P}=\{\mu_k\}$) introduces a critical bottleneck: the sampling operation $\mathbf{P} \sim S(\mathcal{H})$ is non-differentiable.
As a result, the gradients from the reconstruction loss $\mathcal{L}_{rec}$ cannot propagate back to update the heatmap:
\begin{equation}
    \frac{\partial \mathcal{L}_{rec}}{\partial \mathcal{H}} = \frac{\partial \mathcal{L}_{rec}}{\partial \mathbf{P}} \cdot \underbrace{\frac{\partial \mathbf{P}}{\partial \mathcal{H}}}_{\text{gradient detachment}} \approx 0.
\end{equation}
This gradient blockage renders standard end-to-end training infeasible and validates the rationale behind multi-stage learning strategies in InstantGI \cite{InstantGI}. 
To overcome this, we decouple the learning process by mapping the discrete, optimized Gaussian positions back into a continuous density field. 
In other words, we train an individual Gaussian Prior prediction network that approximates the optimal spatial distribution of $K$ Gaussians.

% Since $\Theta$ is unordered, a network outputting a fixed-size tensor faces a permutation ambiguity ($K!$ possible orderings). 
% We observe that while finding the global optimum for Gaussian parameters is ill-posed, finding a reasonable basin of attraction is learnable.
% By enforcing that Gaussian centers $\boldsymbol{\mu}_k$ must be sampled from a spatial probability map $\mathcal{H}$, we anchor the unordered set to the ordered pixel grid. This reduces the search space from a high-dimensional continuous manifold to a structured subspace. The problem transforms from Set Prediction to distribution fitting, which Convolutional Neural Networks (CNNs) can handle naturally.

\subsection{Deep Gaussian Prior}

While previous works often rely on static heuristics: assuming Gaussian positions strictly follow image gradients or edges \cite{ImageGS}, we posit that the optimal spatial distribution is non-stationary and governed by the interplay between local spectral density and global optimization dynamics\cite{Uncertainty}. 

% Inspired by the uncertainty map in super-resolution tasks\cite{Uncertainty}, 
We define the Gaussian Position Prior as a dense field $\mathcal{H}$ (heatmap), and we observed that $\mathcal{H}$ is deeply correlated with the Number of Gaussians ($K$). 
For instance, in a low-bitrate setting (small $K$), Gaussians $\Theta = \{ \mathbf{g}_k \}_{k=1}^K$ must prioritize global coverage to represent the base structure, whereas gradient-based heuristics over-concentrate on edges, leave flat regions underfitting.
Conversely, under high-bitrate settings (large $K$), global structure is initially covered, and the extra points will concentrate on high-frequency details to minimize residual error.
Therefore, we model the heatmap as a conditional distribution $\mathcal{H} \sim P(\mathbf{x} | I, K)$.

We employ a lightweight U-Net architecture for Gaussian Prior Heatmap $\mathcal{H}$ prediction. 
We inject $K$ into the network as a global condition to enable dynamic adaptation to varying compression rates.
Specifically, we normalize the scalar $K$ via logarithmic scaling to handle its large dynamic range:
\begin{equation}
    \hat{k} = \frac{\log(1 + K)}{\log(1 + K_{\max})},
\end{equation}
where $K_{\max}$ is a predefined upper bound (e.g., $10^6$). This normalized scalar is then projected into a high-dimensional embedding $\mathbf{e}_K \in \mathbb{R}^{D}$ via a learnable MLP. 

To modulate the spatial features with $K$, we utilize Feature-wise Linear Modulation (FiLM) at the bottleneck \cite{TemporalFiLM}. Let $\mathbf{F} \in \mathbb{R}^{C \times H \times W}$ denote the bottleneck feature map. We compute channel-wise scale $\boldsymbol{\gamma} \in \mathbb{R}^C$ and shift $\boldsymbol{\beta} \in \mathbb{R}^C$ vectors from $\mathbf{e}_K$ to modulate $\mathbf{F}$ via an affine transformation:
\begin{equation}
    \mathbf{F}'_{c, h, w} = \gamma_c(\mathbf{e}_K) \cdot \mathbf{F}_{c, h, w} + \beta_c(\mathbf{e}_K),
\end{equation}
where the subscripts $c, h, w$ indicate that the modulation parameters are broadcast across spatial dimensions. The decoder mirrors the encoder structure and outputs the final single-channel probability map $\mathcal{H}_{pred}$ or $\hat{\mathcal{H}} \in [0, 1]^{H \times W \times1}$ .
% through a Sigmoid activation function.

% \subsection{Implicit Adversarial Gaussian Initialization}
\subsection{Bootstrapping Gaussian Initialization}

We formulate the Gaussian initialization problem as an iterative bootstrapping learning task: 
We employ an \textbf{Initialization Network} $\mathbf{U}_\omega$ that generates the initial Gaussian distribution by predicting a spatial probability map $\hat{\mathcal{H}}$,
and a \textbf{Global Optimization Process} ${\Phi}$ to optimize the predicted Gaussians subject to reconstruction constraints (e.g., Adam optimization on MSE loss) \cite{Disco2022}.
Let $I$ be the input image. The Initialization Network $\mathbf{U}_\omega$ first predicts a spatial probability distribution $\hat{\mathcal{H}} = \mathbf{U}_\omega(I)$.
A set of $K$ initial Gaussian positions, denoted as  $ \mathbf{P}_{init}$, is then sampled from this predicted heatmap:
\begin{equation}
    \mathbf{P}_{init} \sim S( \mathbf{U}_\omega ({I}), K),
\end{equation}
where $S(\cdot)$ represents the sampling operation. Specifically, we employ a multinomial sampling strategy in this work \cite{MultinomialSamp}.

These initial Gaussians are then processed by the gradient-based optimization process $\Phi(\cdot)$, as implemented in previous 2D Gaussian Splatting works \cite{ImageGS}:
\begin{equation}
    \mathbf{P}_{opt} = \Phi(\mathbf{P}_{init}, {I}).
\end{equation}
In each training loop, the Global Optimizer provides an optimum target $\mathcal{H}_{opt}$. However, since $\mathbf{P}_{opt}$ is a discrete set, we apply a projection operator $\Pi(\cdot)$ to map it back to the continuous domain. The network parameters $\omega$ are updated to match this target:
\begin{align}
   \mathcal{L}_{opt} &= \| \mathcal{H}_{opt} - \hat{\mathcal{H}}\|^2_2, \\
   &= \| \Pi(\mathbf{P}_{opt}) - \mathbf{U}_{\omega }(I)\|^2_2,
\end{align}
While the Initialization Network $\mathbf{U}_\omega$ initially tends to cluster Gaussians excessively around local high-frequency regions \cite{FreeU},
the Global Optimization Process ${\Phi}$ generates strong gradients in under-represented regions, forcibly dispersing the Gaussians to minimize the global reconstruction error.
% $\mathcal{L}_{rec} = \| I - \hat{I} \|$.
Once the prediction of the initialization network aligns with the global optimizer's stationary distribution, convergence is achieved at a balanced point \cite{NashEqGAN}.

% While the Initialization Network $\mathbf{U}_\omega$ tends to over-fit to local high-frequency hotspots (Mode Collapse), clustering Gaussians excessively \cite{}, 
% the Global Optimization Process ${\Phi}$ will generate strong gradients in under-represented regions, forcibly dispersing the Gaussians to minimize the global reconstruction error. 
% Once the prediction of Initialization Network aligns with the Global Optimizer's stationary distribution, resulting in zero displacement of Gaussians, convergence is then reached at a Nash Equilibrium \cite{}.

\subsection{Gaussian Attribute Regression}
\label{sec:feat}
Leveraging the optimized spatial distribution $\mathbf{P}$ derived from the previous stage as a structural prior, we proceed to regress the remaining Gaussian attributes via a separate U-Net.

While Instant-GI typically samples latent feature vectors and decodes via independent MLP process \cite{InstantGI}, we predict a single multi-channel tensor $\mathcal{M} \in \mathbb{R}^{H \times W \times 8}$, which consists of: \textbf{Offset Map} (2 channels) for sub-pixel positional fine-tuning, \textbf{Scale Map} (2 channels) for log-space scales, \textbf{Color Map} (3 channels) for RGB coefficients, and \textbf{Rotation Map} (1 channel) for orientation angles.
This fully convolutional design allows for highly efficient, resolution-independent processing.

We reconstruct the final Gaussian set $\Theta'$ by querying this dense attribute map using the spatial prior $\mathbf{P}$.
Specifically, for each Gaussian center $\boldsymbol{\mu}_k = (u_k, v_k) \in \mathbf{P}$, we sample the corresponding attribute vectors from $\mathcal{M}$ via grid indexing:
\begin{align}
    \boldsymbol{\delta}_k &= \mathcal{M}_{offset}[v_k, u_k], \\ 
    \mathbf{s}_k &= \mathcal{M}_{scale}[v_k, u_k], \\
    \theta_k &= \mathcal{M}_{rot}[v_k, u_k], \\
    \mathbf{c}_k &= \mathcal{M}_{color}[v_k, u_k].
\end{align}
The Gaussian positions are then refined by adding the learned offsets: $\boldsymbol{\mu}'_k = \boldsymbol{\mu}_k + \boldsymbol{\delta}_k$.
Finally, we gather all attributes to form the complete Gaussian set:
\begin{align}
    \Theta' = \{ \mathbf{g}_k \}_{k=1}^K = \{ \boldsymbol{\mu}'_k, \mathbf{s}_k, \mathbf{c}_k, \theta_k  \}_{k=1}^K.
\end{align}
The collected primitives are rendered via the differentiable rasterization process to produce $\hat{I}$. The network is trained end-to-end using the reconstruction loss:
\begin{align}
    \mathcal{L}_{rec} =  \| I - \hat{I} \|^2_2 = \| I - \mathcal{R}(\Theta') \|^2_2.
\end{align}

\begin{figure*}[htbp]
    \centering
    \includegraphics[width=1\linewidth]{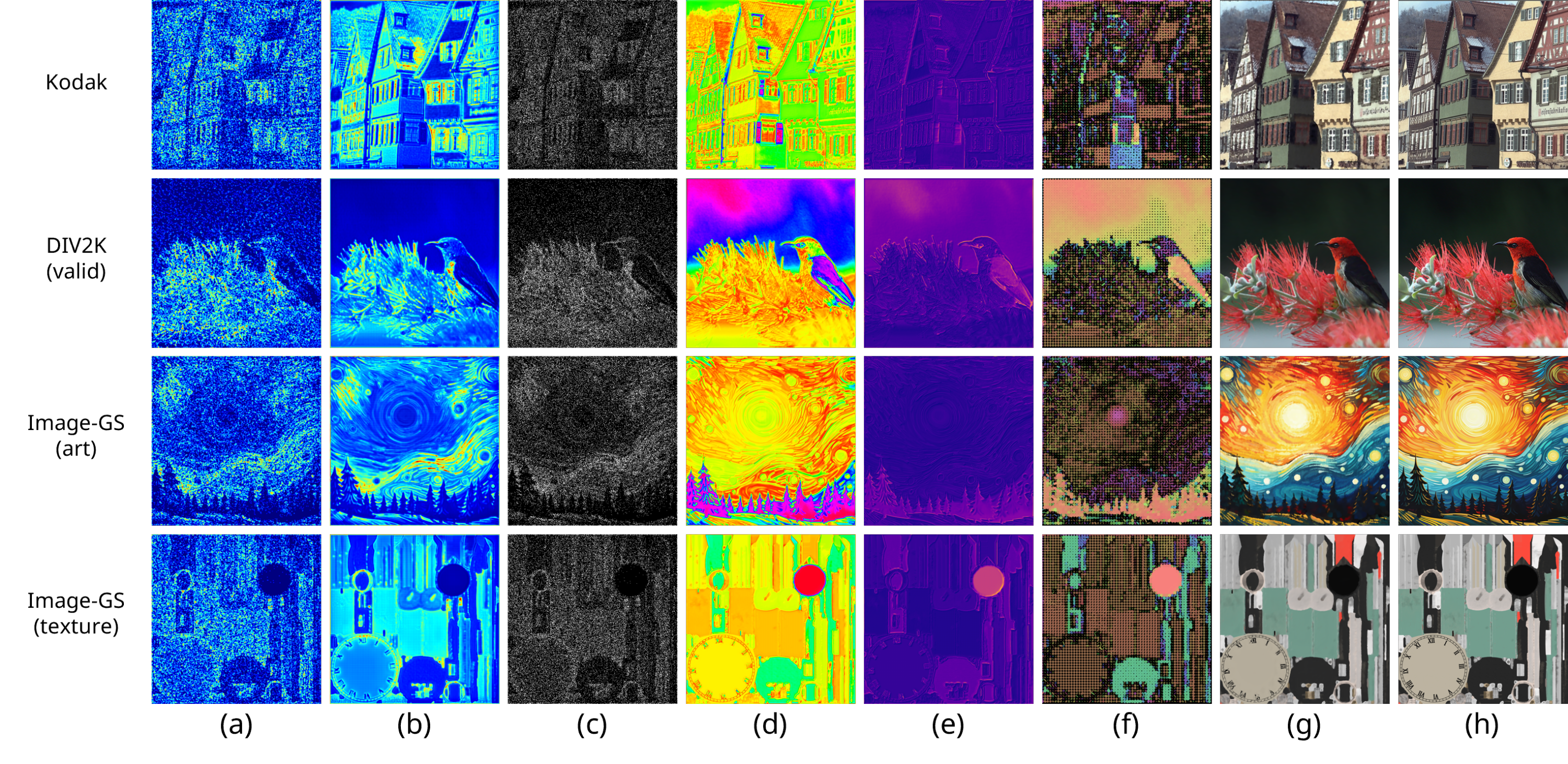}
    \caption{Results of Fast-2DGS. (a) are the reference heatmaps, (b) are the predicted heatmaps from the position network, (c) are the Gaussian positions sampled from the predicted heatmaps, (d) are the feature maps of the last CNN layer, (e) are the offset maps from the attribute network, (f) are the scale-rotation fields, (g) are the predicted Gaussians after rasterization, and (h) are the fine-tuned images.
}
    % \vspace{-0.5cm}
    \label{fig:grid_1}
\end{figure*}

\begin{figure*}[htbp]
    \centering
    \includegraphics[width=1\linewidth]{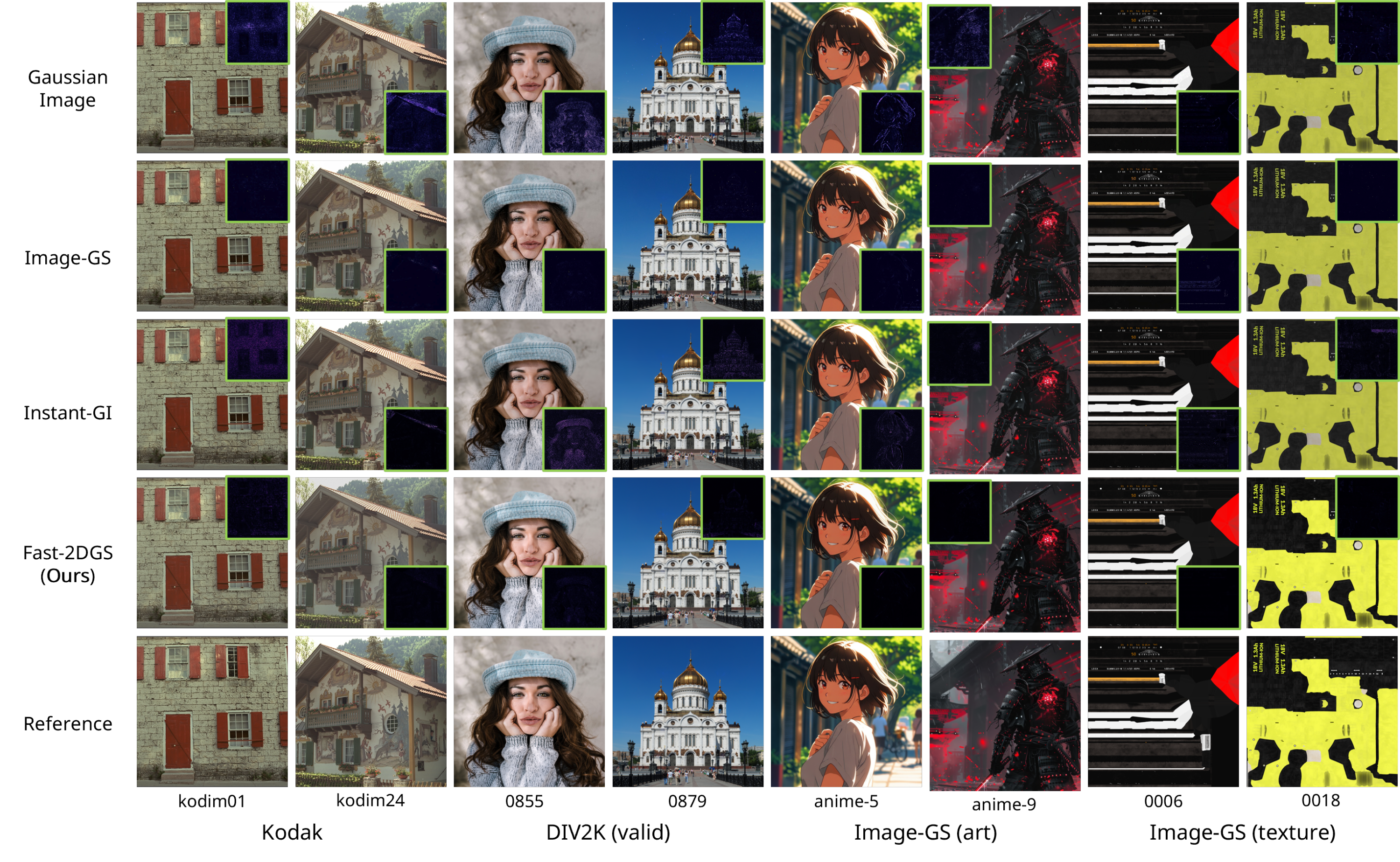}
    \caption{Qualitative comparison against existing Gaussian methods. ROIs represent the error map.}
    % \vspace{-0.5cm}
    \label{fig:grid_2}
\end{figure*}

% ===================================================================

\begin{table*}[htbp]
\centering
\resizebox{1\linewidth}{!}{   
\begin{tabular}{l|ccc|ccc|ccc|ccc|cll|c}
\hline
\multicolumn{1}{c|}{\textbf{Method}}         & \multicolumn{3}{c|}{\textbf{PSNR} $\downarrow$} & \multicolumn{3}{c|}{\textbf{MS-SSIM} $\downarrow$} & \multicolumn{3}{c|}{\textbf{Time (s)} $\downarrow$} & \multicolumn{3}{c|}{\textbf{Training FPS} $\uparrow$} & \multicolumn{3}{c|}{\textbf{Latency} $\downarrow$} & \textbf{Size} $\downarrow$\\ \hline
\multicolumn{1}{c|}{$K$}                     & 10k        & 50k       & 100k      & 10k         & 50k        & 100k       & 10k         & 50k         & 100k       & 10k       & 50k       & 100k      & \multicolumn{3}{c|}{}                 &               \\ \cline{1-13}
GaussianImage (Cholesky)\cite{GaussianImage} & 19.54      & 39.36     & 39.31     & 0.9685      & 0.9968     & 0.9952     & 12.68       & 12.85       & 14.80      & 2852      & 2645      & 2257      & \multicolumn{3}{c|}{-}                & -             \\
GaussianImage (RS)\cite{GaussianImage}       & 30.08      & 39.78     & 34.81     & 0.9684      & 0.9971     & 0.9863     & 13.26       & 13.66       & 15.08      & 2639      & 2459      & 2211      & \multicolumn{3}{c|}{-}                & -             \\
Image-GS (g)~\cite{ImageGS}                  & 33.26      & 39.04     & 39.86     & 0.9777      & 0.9939     & 0.9949     & 27.76       & 27.97       & 30.64      & 1340      & 771       & 508       & \multicolumn{3}{c|}{-}                & -             \\
Image-GS (s)~\cite{ImageGS}                  & 34.12      & 38.65     & 39.73     & 0.9835      & 0.9939     & 0.9949     & 22.91       & 26.48       & 27.86      & 1475      & 777       & 509       & \multicolumn{3}{c|}{-}                & -             \\ \hline
Instant-GI~\cite{InstantGI} ($K$=47,246)& -          & 41.41     & -         & -           & 0.9960     & -          & -           & 10          & -          & \multicolumn{3}{c|}{-}            & \multicolumn{3}{c|}{156.39 ms}         & 1172 MB       \\
\textbf{Fast 2DGS (ours)}                    & 33.79      & 41.85     & 43.01     & 0.9865      & 0.9982     & 0.9988     & 5           & 5           & 5          & \multicolumn{3}{c|}{-}            & \multicolumn{3}{c|}{4.29 ms}           & 29 MB         \\ \hline
\end{tabular}}
\caption{Benchmark of Gaussian image approaches on the Kodak dataset. We round the results of Instant-GI to 50k level for reference.}
\label{tab:bench}
\end{table*}

% \begin{table*}[htbp]
% \centering
% \resizebox{1\linewidth}{!}{   
% \begin{tabular}{l|lll|lll|lll|lll|lll|l}

% \textbf{Method} & \multicolumn{3}{c}{\textbf{PSNR}}& \multicolumn{3}{c}{\textbf{MS-SSIM}}& \multicolumn{3}{c}{\textbf{Time (s)}} & \multicolumn{3}{c}{FPS/Latency} &\multicolumn{3}{c}{Params} &\textbf{Size}\\
%  K& 10k& 50k& 100k&  10k& 50k&100k&  10k& 50k&100k & 10k& 50k&100k  &10k& 50k&100k   & \\

% GaussianImage (Cholesky)\cite{GaussianImage} & 19.54& 39.36&39.31& 0.9685& 0.9968&0.9952& 12.68& 12.85& 14.80& 2852& 2645& 2257&80k& 400k&800k & \\
%  GaussianImage (RS)\cite{GaussianImage}& 30.08& 39.78& 34.81& 0.9684& 0.9971& 0.9863& 13.26& 13.66& 15.08& 2639& 2459& 2211& 80k& 400k& 800k & \\
% Image-GS (g)~\cite{ImageGS} & 33.26& 39.04&39.86& 0.9777& 0.9939&0.9949& 27.76& 27.97& 30.64& 1340& 771& 508&80k& 400k& 800k & \\
% Image-GS (s)~\cite{ImageGS} & 34.12& 38.65& 39.73& 0.9835& 0.9939& 0.9949& 22.91& 26.48& 27.86& 1475& 777& 509& 80k& 400k& 800k & \\ 
% Instant-GI~\cite{InstantGI} & 41.41&&& \multicolumn{3}{c}{0.9960}& \multicolumn{3}{c}{10}& \multicolumn{3}{c}{156.39ms}&& & & \\

% \textbf{Fast 2DGS (ours)} & 33.79& 41.85&43.01& 0.9865& 0.9982&0.9988& \multicolumn{3}{c}{5}& \multicolumn{3}{c}{4.29ms}&& & & \\

% \end{tabular}
% }
% \caption{Benchmark of Gaussian image approaches on Kodak dataset.}
% \label{tab:bench}
% \end{table*}

%-------------------------------------------------------------------------
\section{Experiments}
\label{sec:exp}

\subsection{Implementation Details}
% We implemented our framework using PyTorch and conducted all experiments on a workstation equipped with an Intel Core i9-13900K CPU, 96GB RAM, and a single NVIDIA RTX 4090 GPU (24GB VRAM). 
% Both the Position Network and Attribute Network utilize a lightweight U-Net architecture with a combined model size of merely 29 MB ($\approx$14 MB each), ensuring high deployment efficiency. The Position Network outputs a single-channel probability map, while the Attribute Network predicts an 8-channel tensor encoding offsets, scales, colors, and rotations, as described in Section \label{sec:feat}.
% Training was performed on the DIV2K dataset,
% Although our fully convolutional architecture natively supports arbitrary aspect ratios and resolutions, 
% and we resized and center-cropped all images to $512 \times 512$ during training for benchmarking consistency.
The training is performed on the DIV2K dataset, and the process was divided into two stages using the AdamW optimizer and $\mathcal{L}_2$ loss. 
In \textbf{Stage I}, the Position Network was trained for 200 epochs ($\sim$12 hours) with an initial learning rate of $5 \times 10^{-4}$. Crucially, we randomly sampled the primitive budget $K\sim(10k, 100k)$ during each iteration to enforce complexity-aware initialization.
In \textbf{Stage II}, we froze the Position Network and train the Attribute Network independently for 500 epochs ($\sim$7 hours) with a learning rate of $1 \times 10^{-4}$ to maximize reconstruction fidelity.
Finally, we fine-tune the prediction of rendered images with $5000$ iterations ($\sim 10$ sec) under a learning rate of  $2e-3$ to reach an optimal quality.

\subsection{Datasets}
We evaluate our method on four diverse datasets to ensure comprehensive benchmarking:
\begin{itemize}
    \item \textbf{Kodak PhotoCD Dataset}~\cite{GaussianImage}: A standard benchmark for image reconstruction consisting of 24 lossless, true-color images with a resolution of $768 \times 512$. It is widely used to evaluate compression and restoration algorithms.
    \item \textbf{DIV2K Dataset}~\cite{DIV2K}: A high-quality dataset containing 2K resolution images with diverse real-world content. We utilize the 800 images from the training set for training our models and the 100 images from the validation set for evaluation.
    \item \textbf{Image-GS Stylized Dataset}~\cite{ImageGS}: To test generalization on non-natural statistics, we adopt the dataset collected by Image-GS, which comprises 45 high-resolution ($2048 \times 2048$) images spanning five categories: vector art, photographs, digital art, anime, and oil painting.
    \item \textbf{Image-GS Texture Dataset}~\cite{ImageGS}: This dataset contains 19 texture stacks ($2048 \times 2048$) representing various materials (e.g., wood, fabric, metal). This tests the model's ability to capture high-frequency repetitive patterns crucial for computer graphics workflows.
\end{itemize}

Since image reconstruction quality is jointly determined by resolution and the primitive count $K$, and datasets vary in size, we resized and center-cropped all inputs to a fixed $512 \times 512$. This standardization eliminates resolution as a confounding variable, ensuring a fair assessment of the impact of $K$ across all methods.

% \textbf{Baselines.}
% We compare against three categories of methods:
% \begin{enumerate}
%     \item \textbf{Optimization-based 2DGS:} GaussianImage \cite{gaussianimage} (Random Init) and Image-GS \cite{imagegs} (Gradient Init). These represent the current state-of-the-art for explicit representations.
%     \item \textbf{Network-based 2DGS:} Instant-GI \cite{instantgi}. This is a concurrent work that also attempts network initialization but uses a heavier architecture.
%     \item \textbf{Implicit Neural Representations:} SIREN \cite{sitzmann2020implicit} and WIRE \cite{wire}. These serve as benchmarks for compactness and fitting capacity.
% \end{enumerate}

\subsection{Qualitative Analysis}
Fig. \ref{fig:grid_1} illustrates the overall pipeline of Fast-2DGS. Given input images, the Position Network predicts a Gaussian Position Prior heatmap (Fig. \ref{fig:grid_1}(b)), which is smoother and more continuous than the PPM derived from optimized Gaussian images (Fig. \ref{fig:grid_1}(a)), while still capturing different levels of structural details. 
% This heatmap serves as a reliable prior for Gaussian position sampling.
We then apply Multinomial sampling to select $K$ Gaussian centers from the predicted heatmap (Fig. \ref{fig:grid_1}(c)), enabling effective coverage under limited Gaussian budgets. 
The sampled positions are used to retrieve corresponding attributes from the Attribute Network (Fig. \ref{fig:grid_1}(d)), including offset maps, scale-rotation fields, and color maps. 
The offset and scale-rotation maps provide sub-pixel positional refinement and shape initialization, respectively. 
% while the color map preserves appearance information.
Finally, the predicted Gaussians are rendered by the 2DGS rasterization module to produce an initial reconstruction (Fig. \ref{fig:grid_1}(g)), which is further refined with a few fine-tuning steps to obtain the final high-quality result (Fig. \ref{fig:grid_1}(h)).

Compared with existing methods, we present qualitative comparisons across four datasets. As shown in Fig. \ref{fig:grid_2}, the proposed method achieves visual quality that is comparable to or better than other Gaussian-based approaches, while consistently producing lower reconstruction errors.
In particular, our method effectively suppresses artifacts and blurring, demonstrating improved efficiency and robustness across diverse image content.

% ===================================================================
\begin{table}[htbp]
\centering
\resizebox{1\linewidth}{!}{   
\begin{tabular}{l|cccc}
\hline
\multicolumn{1}{c|}{\textbf{Method}}                           & \multicolumn{4}{c}{\textbf{PSNR} $\downarrow$}                                                                                                                                                                                        \\ \cline{2-5} 
\multicolumn{1}{c|}{($K=50,000$)}                                & \textbf{Kodak}                            & \textbf{\begin{tabular}[c]{@{}c@{}}DIV2K\\ (valid)\end{tabular}} & \textbf{\begin{tabular}[c]{@{}c@{}}ImageGS\\ (art)\end{tabular}} & \textbf{\begin{tabular}[c]{@{}c@{}}ImageGS\\ (texture)\end{tabular}} \\ \hline
GaussianImage (Cholesky) \cite{GaussianImage} & 28.19                                     & 20.91                                                            & 28.15                                                            & 29.09                                                                \\
GaussianImage (RS) \cite{GaussianImage}       & 29.20                                     & 28.39                                                            & 28.07                                                            & 28.90                                                                \\
Image-GS (G) \cite{ImageGS}                   & 31.12                                     & 27.12                                                            & 29.21                                                            & 31.27                                                                \\
Image-GS (S) \cite{ImageGS}                   & 30.49                                     & 26.80                                                            & 27.25                                                            & 29.15                                                                \\
Instant-GI \cite{InstantGI}                   & \cellcolor{lime!30}37.89 & \cellcolor{lime!70}38.01                        & \cellcolor{lime!30}40.38                        & \cellcolor{lime!30}41.69                            \\ \hline
Fast 2DGS \textbf{(ours)}                     & \cellcolor{lime!70}38.14 & \cellcolor{lime!30}37.81                        & \cellcolor{lime!70}41.21                        & \cellcolor{lime!70}42.51                            \\ \hline
\end{tabular}}
\caption{Reconstruction performance comparison across all datasets. Specifically, $K$=50,000, and PSNR is calculated at $1st$ second of fine-tuning.}
\label{tab:datasets}
\end{table}

\subsection{Quantitative Results}

We primarily evaluate reconstruction quality using Peak Signal-to-Noise Ratio (PSNR) and Multi-Scale Structural Similarity Index (MS-SSIM), where PSNR measures pixel-level reconstruction fidelity, and MS-SSIM captures perceptual and structural similarity across multiple scales \cite{MS-SSIM}.
% Table \ref{tab:bench} reports quantitative comparisons with Gaussian budgets $K=10{,}000$, $50{,}000$, and $100{,}000$. Optimization-based methods, including GaussianImage and Image-GS, achieve reasonable PSNR but require significantly longer optimization time and exhibit a clear quality saturation around 40 dB. Instant-GI attains higher reconstruction quality but lacks flexible control over the Gaussian budget, making the compression rate difficult to adjust. In contrast, our method consistently achieves high reconstruction quality with fast runtime while allowing explicit control over $K$, providing a better trade-off between quality, efficiency, and flexibility.
Table \ref{tab:bench} reports quantitative comparisons with $K=10{,}000$, $50{,}000$, and $100{,}000$. 
In GaussianImage, two covariance parameterizations are considered: Cholesky decomposition and rotation-scaling (RS) decomposition\cite{GaussianImage}. They are equivalent in representation capacity but differ in optimization and compression robustness. While Cholesky performs poorly at low $K$, the RS method faces degradation when $K$ is redundant.
For Image-GS, the reported variants (g) and (s) correspond to gradient-based and saliency-based initialization, respectively, representing two heuristic strategies for Gaussian placement. Compared to GaussianImage, Image-GS's content-adaptive design achieves higher PSNR with more Gaussians.
However, both optimization-based methods, while achieving decent PSNR, require longer optimization time and exhibit a clear quality saturation around 40 dB.
Instant-GI attains higher reconstruction quality but lacks flexible control over the Gaussian budget, making the compression rate difficult to adjust. In contrast, our method consistently achieves high reconstruction quality with fast runtime while allowing explicit control over $K$, providing a better trade-off between quality, efficiency, and flexibility.

\begin{table}[htbp]
\centering
\resizebox{\linewidth}{!}{
\begin{tabular}{l|cccccc|c}
\hline
\multicolumn{1}{c|}{\textbf{Method}} & \multicolumn{6}{c|}{\textbf{PSNR} $\downarrow$}                                                                                                                 & Total \\ \cline{2-7}
\multicolumn{1}{c|}{($K=50,000$)}      & Init.                    & 1s                       & 2s                       & 5s                       & 10s                      & Final                    & time  \\ \hline
G.Image (Chy.)\cite{GaussianImage}   & 13.42                    & 28.19                    & 34.96                    & 38.28                    & 39.20                    & 39.36                    & 13s   \\
G.Image (RS)\cite{GaussianImage}     & 13.68                    & 29.31                    & 35.82                    & 38.86                    & 39.62                    & 39.78                    & 14s   \\
Image-GS (g)\cite{ImageGS}           & 27.80                    & 31.12                    & 33.08                    & 34.12                    & 36.19                    & 39.04                    & 28s   \\
Image-GS (S)\cite{ImageGS}           & 27.53                    & 30.49                    & 30.87                    & 33.14                    & 35.78                    & 38.65                    & 28s   \\
Instant-GI\cite{InstantGI}           & \cellcolor{lime!70}28.26 & \cellcolor{lime!30}37.89 & \cellcolor{lime!30}39.27 & \cellcolor{lime!30}40.60 & \cellcolor{lime!30}41.41 & \cellcolor{lime!30}41.41 & 10s   \\ \hline
Fast 2DGS \textbf{(ours)}            & \cellcolor{lime!30}28.07 & \cellcolor{lime!70}38.14 & \cellcolor{lime!70}40.05 & \cellcolor{lime!70}41.85 & \cellcolor{lime!70}43.13 & \cellcolor{lime!70}43.13 & 10s   \\
(w/o Positions)                        & 14.21                    & 31.14                    & 33.50                    & 36.06                    & 39.96                    & 39.96                    & 10s   \\
(w/o Attributes)                       & 10.37                    & 25.07                    & 28.62                    & 32.28                    & 35.69                    & 35.69                    & 10s   \\
(w/o Both)                             & 9.29                     & 27.61                    & 30.07                    & 33.85                    & 35.55                    & 35.55                    & 10s   \\ \hline
\end{tabular}}
\caption{Reconstruction performance on Kodak (average over 24 images). $K$=50,000. For Instant-GI, $K$=47,246).}
\label{tab:sec}
\end{table}

% Meanwhile, we have tested all methods across 4 datasets for generalizability. Specifically, we use the PSNR of $1$ second fine-tuning as the results. As shown in Table \ref{tab:datasets}, our method maintained 37-38 PSNR on natural images and over 40 PSNR on stylistic arts and textures. 

Meanwhile, we evaluate the generalizability of all methods. Specifically, we report the PSNR after 1 second of fine-tuning for all methods. As shown in Table \ref{tab:datasets}, our method consistently achieves 37–38 dB PSNR on natural images and exceeds 40 dB PSNR on stylized images and texture datasets.

We further evaluate the convergence speed of different methods during Gaussian image optimization on the Kodak dataset. As shown in Table \ref{tab:sec}, our method reaches 40 dB PSNR within 2 seconds, whereas other methods require 5 seconds or more to achieve comparable quality. Moreover, our method attains a higher final PSNR, highlighting the importance of effective Gaussian initialization.

% \begin{table}[h]
% \centering
% \resizebox{\linewidth}{!}{
% \begin{tabular}{l|cccccc|c}
% \toprule
% \multicolumn{1}{c|}{\textbf{Method}} &  \multicolumn{6}{c}{\textbf{PSNR} $\downarrow$}& Total\\
%   ($K=50,000$)&  Init.&1s& 2s& 5s &10s&  Final& time\\ \midrule
%  G.Image (Chy.)\cite{GaussianImage}& 13.42& 28.19& 34.96& 38.28& 39.20&  39.36& 13s\\
%  G.Image (RS)\cite{GaussianImage}& 13.68& 29.31& 35.82& 38.86& 39.62&  39.78& 14s\\
%  Image-GS (g)\cite{ImageGS} & 28.80& 31.12& 33.08& 34.12& 36.19&  39.04& 28s\\
%  Image-GS (S)\cite{ImageGS}& 27.53& 30.49& 30.87& 33.14& 35.78&  38.65& 28s\\
% Instant-GI\cite{InstantGI} &  28.26&37.89& 39.27&  40.60&41.41&  41.41& 10s\\
% \midrule
% Fast 2DGS \textbf{(ours)}&  28.07&38.14& 40.05&  41.85&43.13&  43.13& 10s\\
% \bottomrule
% \end{tabular}
% }
% \caption{Reconstruction performance on Kodak (average over 24 images). $K$=50,000. For Instant-GI, $K$ = 47,246).}
% \label{tab:sec}
% \end{table}

% ===================================================================

\begin{figure}[htbp]
    \centering
    \includegraphics[width=1\linewidth]{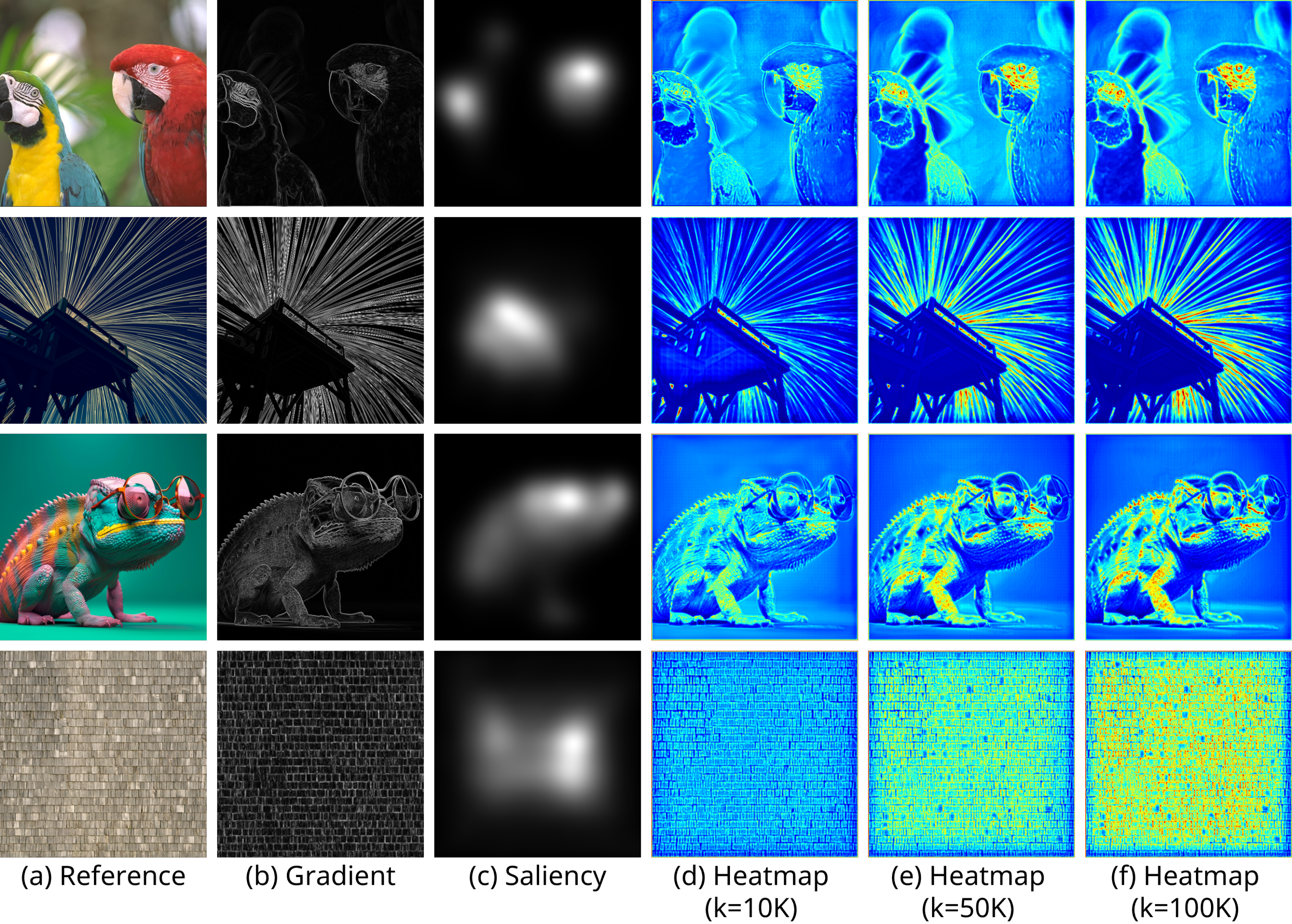}
    \caption{Gaussian initialization comparison. (a) RGB input, (b) Gradient init. (c) Saliency init. (d)(e)(f) Heatmap under $K$ Gaussians.}
    \vspace{-0.5cm}
    \label{fig:heat_1}
\end{figure}

\subsection{Gaussian Prior Evaluation}
To demonstrate the complexity-awareness of our method, we evaluate its behavior under different $K$.
As shown in Fig. \ref{fig:heat_1}, the predicted Gaussian Prior heatmap $\mathcal{H}_K$ adapts to different values of $K$ accordingly.
When Gaussians are limited, the network encourages a uniform initialization across the image, prioritizing global structure coverage. 
As $K$ increases, the network concentrates probability mass on more informative regions, allocating additional Gaussians to detail-rich areas. 
Compared with heuristic initializations used in Image-GS. The gradient-based initialization, while intuitive, may fail to adequately cover low-frequency areas. 
The saliency-based initialization successfully captures semantic regions, yet often assigns redundant Gaussians to homogeneous regions. 
% Our Gaussian prior effectively addresses both limitations by jointly considering global structure and local complexity, resulting in a more balanced and budget-aware Gaussian allocation.
% Our adaptive design addresses both limitations effectively by automatically balancing global coverage and local detail refinement according to the available representation units.

% To demonstrate the complexity-awareness, we test the Position Network performance under different $K$. Specifically, we display the Gaussian Prior Heatmap $\mathcal{H}_K$. As shown in Figure \ref{fig:heat_1}, the predicted Gaussian Prior Heatmap varing under different settings. Specifically, when number of Gaussians is under 10,000, the Heatmap tend to guide the Gaussians initialize uniformly across images. When $K$ is increasing, the Gaussians will be assigned to cover more areas, as shown in Figure \ref{fig:heat_1} (). Therefore, additional Gaussians will be assigned to key areas to enhance details, as shown in Figure \ref{fig:heat_1} ().

% ===================================================================

\subsection{Performance and Ablation Studies}

\noindent \textbf{Ablation Analysis}
We conduct ablation experiments to analyze the impact of different components in Gaussian initialization, as show in Table \ref{tab:sec}. 
When removing the position prediction, the initial reconstruction quality drops to 14 dB PSNR. Nevertheless, due to the preserved attribute initialization, the model is able to recover to 39 dB PSNR after 10 seconds of fine-tuning, still outperforming optimization-based methods.
In contrast, when replacing the attributes with random initialization, the initial PSNR further degrades to 10 dB, and the final performance saturates at around 35 dB PSNR even after 10 seconds of optimization. 
Removing both position and attribute predictions reduces the method to random Gaussian initialization
% , resulting in performance comparable to Image-GS with random initialization, as both share the same rasterization kernel.
% These results demonstrate that Gaussian attribute initialization plays a critical role in determining overall image quality, while Gaussian position initialization is the primary factor limiting the achievable final PSNR. In particular, accurate initial placement of Gaussians enables faster convergence and higher-quality reconstruction during subsequent optimization. This ablation study directly validates the effectiveness of our proposed Gaussian prior–based position prediction.

\begin{figure}[htbp]
    \centering
    \includegraphics[width=1\linewidth]{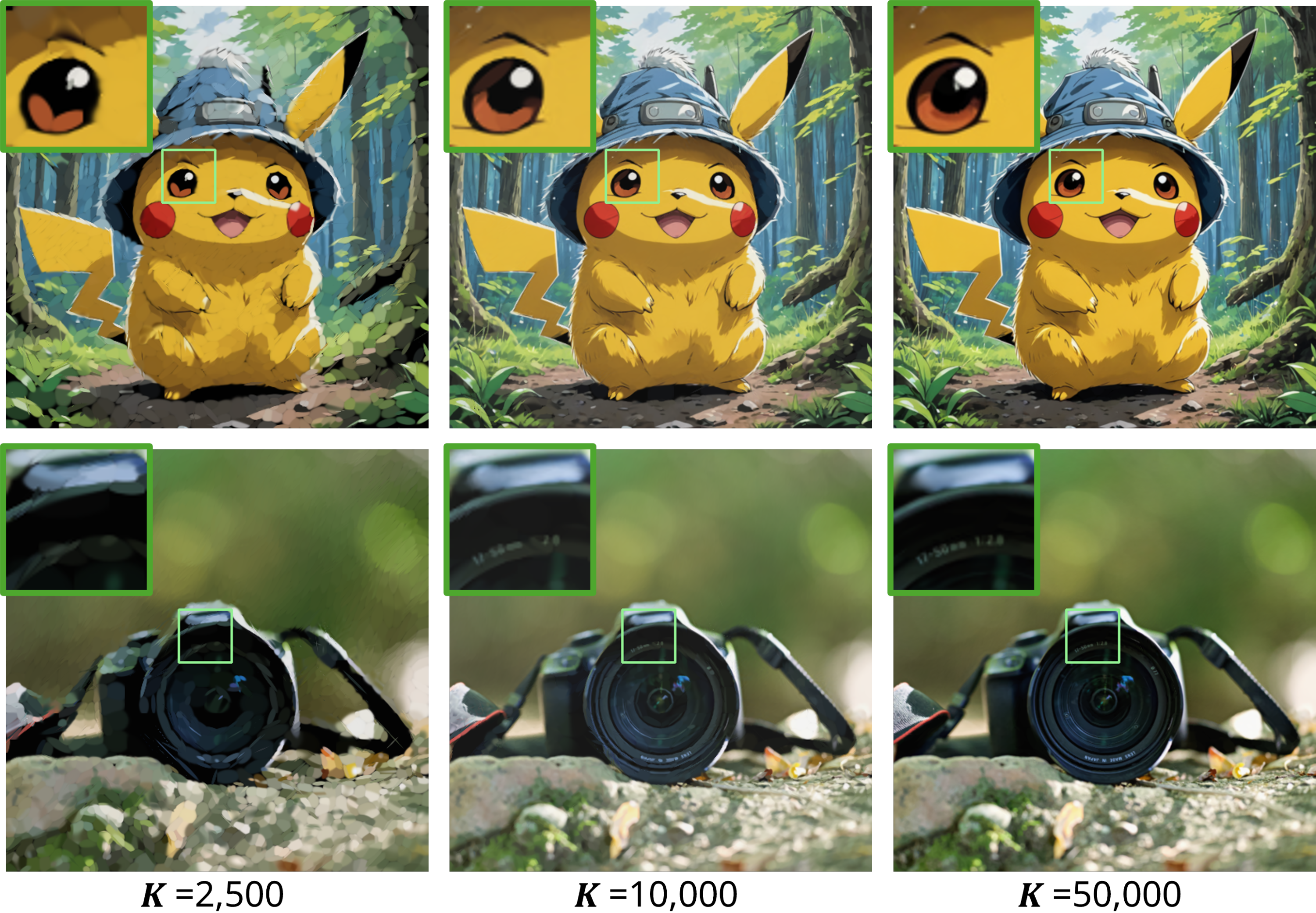}
    \caption{ROIs (Region of Interest) under different $K$.}
    \vspace{-0.5cm}
    \label{fig:roi}
\end{figure}

\noindent \textbf{Compression Analysis}
As shown in Fig. \ref{fig:roi}, our method supports flexible compression by adjusting $K$ to arbitrary values.
Specifically, with $K=50,000$, the reconstructed image achieves lossless quality, where fine details and textures remain clearly preserved. 
When $K$ is reduced to $10,000$, minor detail loss can be observed, yet the overall visual quality remains high and perceptually faithful. 
Under more aggressive compression ($K=2{,}500$), fine-grained details, particularly structured textures such as text, begin to vanish; however, the global semantic structure is largely preserved, and no severe blurring or smearing artifacts are introduced. 
These results demonstrate that our representation degrades gracefully under compression, retaining semantic integrity even at very limited Gaussian primitives.

\noindent \textbf{Computation Analysis}
Both our method and Instant-GI adopt network predictions followed by a fine-tuning process. 
In terms of network inference latency, our method requires only 4.29 ms, whereas Instant-GI incurs a significantly higher cost of 156.39 ms, as shown in Table \ref{tab:bench}. 
Moreover, Instant-GI relies on a heavy network with a weight size of 1,172 MB, while our two lightweight networks together occupy only 29 MB in total (14.68 MB + 14.82 MB). 
This substantial reduction in latency and model size makes our design more suitable for practical tasks.

\section{Conclusion}
\label{sec:conclusion}

% We have presented \methodname, a paradigm shift in 2D Gaussian Splatting for image representation. By framing initialization as a "learning to optimize" problem, we successfully distilled the expensive iterative optimization process into a lightweight, feed-forward network. Our Deep Gaussian Prior, combined with a decoupled attribute regression stage, solves the initialization bottleneck, enabling high-fidelity reconstruction in sub-second times. 

We conclude that the training of Gaussian images is heavily dependent on initialization, and we demonstrate that initialization priors can be effectively modeled by deep learning. Experiments on four representative datasets show that it requires no extensive feature engineering and heavy computation resources to achieve state-of-the-art performance, but a simple neural network.
However, while we believe our approach paves the way for the next generation of image codecs, this is only the inception of Gaussian image representation. 
Despite low latency, our current network design may bottleneck performance on high-resolution images (e.g., 4K/8K). 
That said, Gaussian image representation still has a long way to meet industry standards for real-time, end-to-end compression required in the AIGC era. 
Therefore, our future work will explore scalable learning and explicit feature learning in generative models to close this gap.

% Derived by our experimet reuslts, we conclude that Gaussian Image training process is deeply corelated with the initialization of its Gaussians. Furthermore, we prove that Gaussian initialization priors can be effectively modeled by deep learning method. Tested on four representive dataset, we demonstrated that achieving state-of-the-art level of perfoamnce requires no extenisve feature engineering but simple neural networks, and the computation cost can be significantly reduced without perfomance trade-off.
% However, these approaches are only the start of the Gaussian image representation. Our method despite low latency, is still far from the industry level of image compression which required real-time end-to-end encoding and decoding performance in AIGC era. Thus, scalable learning and explicit feature learning in generative models pose a direction of our future work.

%-------------------------------------------------------------------------

{
    \small
    \bibliographystyle{ieeenat_fullname}
    \bibliography{main}
}

\end{document}